\documentclass[journal]{IEEEtran}
\usepackage[utf8]{inputenc} % allow utf-8 input
\usepackage[T1]{fontenc}    % use 8-bit T1 fonts
\usepackage{hyperref}       % hyperlinks
\usepackage{url}            % simple URL typesetting
\usepackage{booktabs}       % professional-quality tables
\usepackage{amsfonts}       % blackboard math symbols
\usepackage{nicefrac}       % compact symbols for 1/2, etc.
\usepackage{microtype}      % microtypography
\usepackage{xcolor}         % colors
\usepackage[pdftex]{graphicx}
\usepackage{tabularx}
\usepackage{arydshln}
\usepackage{amsmath}
\usepackage{amsthm}
\usepackage{xcolor}
\usepackage{multicol}
\usepackage{multirow}
\usepackage{xspace}
\usepackage[caption=false,font=normalsize,labelfont=sf,textfont=sf]{subfig}

\definecolor{red}{rgb}{0.99, 0.02, 0.02}

\NewDocumentCommand{\heng}
{ mO{} }{\textcolor{red}{\textsuperscript{\textit{Heng}}\textsf{\textbf{\small[#1]}}}}

\NewDocumentCommand{\qingyun}
{ mO{} }{\textcolor{orange}{\textsuperscript{\textit{Qingyun}}\textsf{\textbf{\small[#1]}}}}

\NewDocumentCommand{\kexuan}
{ mO{} }{\textcolor{green}{\textsuperscript{\textit{Kexuan}}\textsf{\textbf{\small[#1]}}}}

\NewDocumentCommand{\junyu}
{ mO{} }{\textcolor{blue}{\textsuperscript{\textit{Junyu}}\textsf{\textbf{\small[#1]}}}}
\theoremstyle{definition} % Default style for definitions
\newtheorem{definition}{Definition}

\usepackage{cite}
\usepackage{array}

\begin{document}
%
% paper title
% Titles are generally capitalized except for words such as a, an, and, as,
% at, but, by, for, in, nor, of, on, or, the, to and up, which are usually
% not capitalized unless they are the first or last word of the title.
% Linebreaks \\ can be used within to get better formatting as desired.
% Do not put math or special symbols in the title.
\title{
Gene-Metabolite Association Prediction with Interactive Knowledge Transfer Enhanced Graph for Metabolite Production
}

% Gene-Protein Reaction
%
%
% author names and IEEE memberships
% note positions of commas and nonbreaking spaces ( ~ ) LaTeX will not break
% a structure at a ~ so this keeps an author's name from being broken across
% two lines.
% use \thanks{} to gain access to the first footnote area
% a separate \thanks must be used for each paragraph as LaTeX2e's \thanks
% was not built to handle multiple paragraphs
%

\author{Kexuan Xin, Qingyun Wang, Junyu Chen, Pengfei Yu, Huimin Zhao, Heng Ji \\
University of Illinois at Urbana-Champaign\\
xkxjane@gmail.com
        % John~Doe,~\IEEEmembership{Fellow,~OSA,}
        % and~Jane~Doe,~\IEEEmembership{Life~Fellow,~IEEE}% <-this % stops a space
% \thanks{M. Shell was with the Department
% of Electrical and Computer Engineering, Georgia Institute of Technology, Atlanta,
% GA, 30332 USA e-mail: (see http://www.michaelshell.org/contact.html).}% <-this % stops a space
% \thanks{J. Doe and J. Doe are with Anonymous University.}% <-this % stops a space
% \thanks{Manuscript received April 19, 2005; revised August 26, 2015.}
}

% note the % following the last \IEEEmembership and also \thanks - 
% these prevent an unwanted space from occurring between the last author name
% and the end of the author line. i.e., if you had this:
% 
% \author{....lastname \thanks{...} \thanks{...} }
%                     ^------------^------------^----Do not want these spaces!
%
% a space would be appended to the last name and could cause every name on that
% line to be shifted left slightly. This is one of those "LaTeX things". For
% instance, "\textbf{A} \textbf{B}" will typeset as "A B" not "AB". To get
% "AB" then you have to do: "\textbf{A}\textbf{B}"
% \thanks is no different in this regard, so shield the last } of each \thanks
% that ends a line with a % and do not let a space in before the next \thanks.
% Spaces after \IEEEmembership other than the last one are OK (and needed) as
% you are supposed to have spaces between the names. For what it is worth,
% this is a minor point as most people would not even notice if the said evil
% space somehow managed to creep in.

% The paper headers
\markboth{Journal of \LaTeX\ Class Files,~Vol.~14, No.~8, August~2015}%
{Shell \MakeLowercase{\textit{et al.}}: Bare Demo of IEEEtran.cls for IEEE Journals}
% The only time the second header will appear is for the odd numbered pages
% after the title page when using the twoside option.
% 
% *** Note that you probably will NOT want to include the author's ***
% *** name in the headers of peer review papers.                   ***
% You can use \ifCLASSOPTIONpeerreview for conditional compilation here if
% you desire.

% If you want to put a publisher's ID mark on the page you can do it like
% this:
%\IEEEpubid{0000--0000/00\$00.00~\copyright~2015 IEEE}
% Remember, if you use this you must call \IEEEpubidadjcol in the second
% column for its text to clear the IEEEpubid mark.

% use for special paper notices
%\IEEEspecialpapernotice{(Invited Paper)}
\newcommand{\modelname}{IKT4Meta\xspace}

% make the title area
\maketitle

% As a general rule, do not put math, special symbols or citations
% in the abstract or keywords.

\begin{abstract}
In the rapidly evolving field of metabolic engineering, the quest for efficient and precise gene target identification for metabolite production enhancement presents significant challenges. 
Traditional approaches, whether knowledge-based or model-based, are notably time-consuming and labor-intensive, due to the vast scale of research literature and the approximation nature of genome-scale metabolic model (GEM) simulations.
 Therefore, we propose a new task,
 % Gene-Protein Reaction(GPR) Association Prediction
 Gene-Metabolite Association Prediction
 based on metabolic graphs, to automate the process of candidate gene discovery for a given pair of metabolite and candidate-associated genes, as well as presenting the first benchmark containing 2474 metabolites and 1947 genes of two commonly used microorganisms \textit{Saccharomyces cerevisiae (SC)} and \textit{Issatchenkia orientalis (IO)}. 
 This task is challenging due to the incompleteness of the metabolic graphs and the heterogeneity among distinct metabolisms.
 % Specifically,  given a pair of metabolite and candidate-associated genes, our framework can identify the feasible genes effectively.
 % However, performing association prediction on metabolic graphs suffers from the incompleteness of graph structure. Moreover, mitigating the incompleteness issue by integrating knowledge from multiple metabolism graphs is hindered by the heterogeneity of distinct metabolisms.
 To overcome these limitations, we propose an \textbf{I}nteractive \textbf{K}nowledge \textbf{T}ransfer mechanism based on \textbf{M}etabolism \textbf{G}raph (\modelname), which improves the association prediction accuracy by integrating the knowledge from different metabolism graphs. 
 First, to build a bridge between two graphs for knowledge transfer, we utilize Pretrained Language Models (PLMs) with external knowledge of genes and metabolites to help generate inter-graph links, significantly alleviating the impact of heterogeneity.
 Second, we propagate intra-graph links from different metabolic graphs using inter-graph links as anchors.
 Finally, we conduct the gene-metabolite association prediction based on the enriched metabolism graphs, which integrate the knowledge from multiple microorganisms.
 Experiments on both types of organisms demonstrate that our proposed methodology outperforms baselines by up to 12.3\% across various link prediction frameworks. 

\end{abstract}


% For peer review papers, you can put extra information on the cover
% page as needed:
% \ifCLASSOPTIONpeerreview
% \begin{center} \bfseries EDICS Category: 3-BBND \end{center}
% \fi
%
% For peerreview papers, this IEEEtran command inserts a page break and
% creates the second title. It will be ignored for other modes.
\IEEEpeerreviewmaketitle

% The very first letter is a 2 line initial drop letter followed
% by the rest of the first word in caps.
% 
% form to use if the first word consists of a single letter:
% \IEEEPARstart{A}{demo} file is ....
% 
% form to use if you need the single drop letter followed by
% normal text (unknown if ever used by the IEEE):
% \IEEEPARstart{A}{}demo file is ....
% 
% Some journals put the first two words in caps:
% \IEEEPARstart{T}{his demo} file is ....
% 
% Here we have the typical use of a "T" for an initial drop letter
% and "HIS" in caps to complete the first word.

\section{Introduction}

% \qingyun{need to reorganize}
\IEEEPARstart{M}{etabolic} engineering harbors the potential to engineer microorganisms into efficient cell factories for producing value-added products in an economical way, offering significant potential for advancements in medicine, agriculture, and bio-energy~\cite{volkMetabolicEngineeringMethodologies2023}. To achieve this, researchers increase metabolite production by overexpressing the upstream pathway genes or downregulating and knocking out the genes of competitive pathways~\cite{boobEnablingPathwayDesign2024}. Traditional manual pipelines for identifying gene targets include two types: the knowledge-based approach, which relies on human experts to extract relevant information and propose target genes from previous research papers~\cite{boobEnablingPathwayDesign2024}; and the model-based approach, which involves the curation and simulation of Genome-scale metabolic models (GEMs)~\cite{burgardOptknockBilevelProgramming2003,ranganathanOptForceOptimizationProcedure2010,o2015using} to generate candidate genes. However, those manual pipelines have proved time-consuming and laborious~\cite{volkMetabolicEngineeringMethodologies2023}, taking several months to years to complete. 
Moreover, many GEMs contain knowledge gaps, hindering their potential for predictive accuracy and usability \cite{orthSystematizingGenerationMissing2010}.
For example, the current  YeastGEM 9.0.0 for \textit{Saccharomyces cerevisiae} contains only 1,162 genes \cite{zhangYeast9ConsensusYeast2023a}, while 6,617 open reading frames are supposed to encode genes in Saccharomyces Genome Database~\cite{cherry1998sgd}. Additionally, 2,476 genes are still annotated as unknown regarding their molecular function \cite{wongSaccharomycesGenomeDatabase2023}. Therefore, scientists propose gap-filling methods to complete those knowledge gaps, through either topology~\cite{satish2007optimization,thiele2014fastgapfill,durotGenomescaleModelsBacterial2009} or machine-learning methods~\cite{chenTeasingOutMissing2023}.

\begin{figure}[htb!]
\centering
\vspace{-2em}
  \includegraphics[width=\columnwidth]{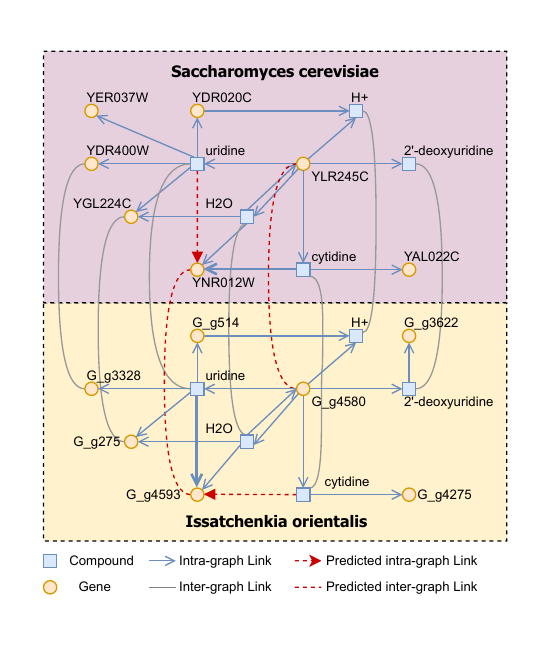}
\vspace{-3em}
  \caption{An example of the subgraphs from Saccharomyces cerevisiae (SC) and Issatchenkia orientalis (IO). Each metabolic graph contains genes and metabolites, with intra-graph links showing gene-metabolite associations and inter-graph links connecting equivalent genes or metabolites. Predictions for these links are based on existing associations. For example, (\texttt{cytidine} $\rightarrow$ \texttt{G\_g514}) shows that \texttt{G\_g514} is involved in a reaction with \texttt{cytidine} as a reactant, while (\texttt{YLR245C} $\rightarrow$ \texttt{H}$^+$) indicates \texttt{YLR245C} catalyzes a reaction producing \texttt{H}$^+$. 
  % \qingyun{need to add explanation caption, change network to graph}
  }
  \label{fig:walk}
\end{figure}

Despite this, previous gap-filling processes mainly focus on identifying the missing reactions for dead-end metabolites that cannot be produced or consumed. Moreover, no universally applicable method yet exists for predicting unknown reactions, particularly in non-model organisms. To address these bottlenecks, we propose a new task, gene-metabolite association prediction, replacing the laborious literature mining and database analysis with gene-metabolite association inference based on existing incomplete metabolic graphs, as shown in Figure~\ref{fig:walk}. 
Specifically, given a potential gene-metabolite link in incomplete metabolic graphs, the model will predict whether such a link is valid. 

However, gene-metabolite association prediction based on metabolic graphs faces two challenges: \textit{incomplete metabolic graphs} and \textit{disparity between different microorganisms}. One major challenge is the unreliability of association predictions based solely on underexplored metabolic graph structures. For example, approximately 10\%-20\% of the genes from Issatchenkia orientalis have no information or have only ambiguous annotations. 
Inspired by the knowledge fusion tasks that transfer knowledge from well-studied objects to under-studied ones~\cite{hinton2015distilling, liu-etal-2024-named,luo2024knowla}, or integrate knowledge from multiple sources \cite{wang-etal-2020-reviewrobot,sun2023joint, sun2023makes,wang-etal-2024-scimon}, we can enrich the metabolic graphs by integrating knowledge across various metabolic graphs. 
Nevertheless, integrating graphs requires inter-graph links, which are a pair of vertices that are identical or highly similar, to connect two graphs for knowledge transfer. 
Detecting such inter-graph links is difficult due to the disparity between different microorganisms and the heterogeneous structure between their metabolic graphs. 
Meanwhile, vertices that fail to connect with counterparts in other graphs remain isolated and are referred to as ``\textit{danglings}'' \cite{sun2021knowing}.
The existence of ``\textit{danglings}'' severely hinders the discovery of inter-graph links. 

Figure \ref{fig:walk} demonstrates the heterogeneity between two metabolic subgraphs in \textit{Saccharomyces cerevisiae} (SC)~\cite{luConsensusCerevisiaeMetabolic2019b}, \textit{Issatchenkia orientalis} (IO)~\cite{suthersGenomescaleMetabolicReconstruction2020}. For instance, the \texttt{uridine} in two microorganisms has different neighbors. Additionally, the presence of ``\textit{danglings}'' complicates the inference of inter-graph links. For example, \texttt{YDR020C} and \texttt{G\_g514} are different 2-hop neighbors of the inter-graph link (\texttt{YLR245C}, \texttt{G\_g4580}). Removing these ``\textit{danglings}'' would greatly simplify the discovery of inter-graph links. 
Moreover, identifying additional inter-graph links can help complete missing gene-metabolite associations.
For instance, by resolving the inter-graph links (\texttt{YNR012W}, \texttt{G\_4593}), it becomes straightforward to establish the intra-graph links from (\texttt{uridine} to \texttt{YNR012W}) and from (\texttt{cytidine} to \texttt{G\_G4593}). 
 Therefore, our primary task is to identify the inter-graph links and use them to enrich the intra-graph links. 

\begin{figure}
  \centering
  \includegraphics[width=\columnwidth]{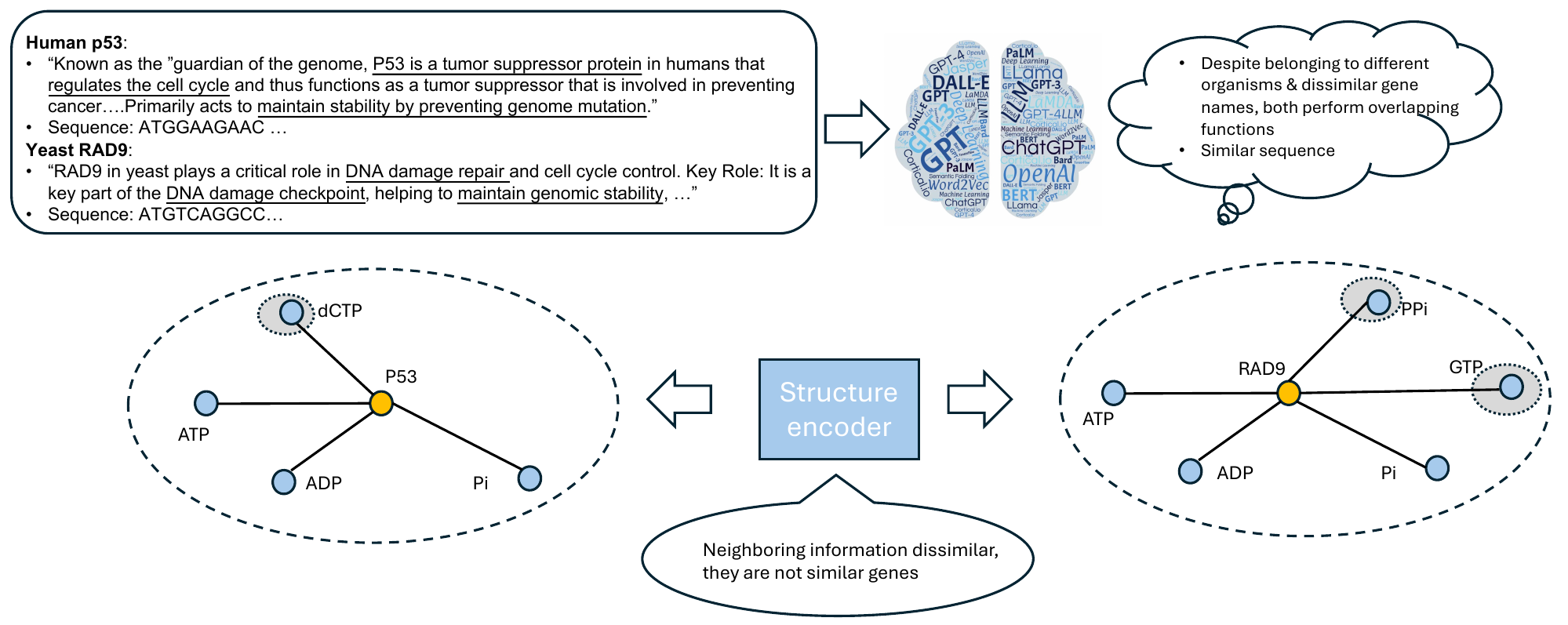}
  \caption{Comparison between text and graph structures for gene alignment. The ScholarBERT \cite{hong2023diminishing} successfully identifies that \texttt{P53} and \texttt{RAD9} from different microorganisms are highly similar based on textual information. However, due to the significant differences in structural context between them, the structure encoder~\cite{kipf2016semi} fails to recognize this pair of similar genes.}
  \label{fig:comparison}
\vspace{-1em}
\end{figure}

To achieve this goal, we propose a new \textbf{I}nteractive \textbf{K}nowledge \textbf{T}ransfer on \textbf{M}etabolism \textbf{G}raph  (\modelname) framework with a graph enrichment stage and an association prediction stage. In the graph enrichment stage, we capture external knowledge of genes and metabolites and utilize structural and relational learning through structure encoders to understand patterns across different graphs. Specifically, due to the complementarity of text descriptions and graph structures (Figure~\ref{fig:comparison}), we propose a multi-modal encoder that leverages Pre-trained language models (PLMs) to process textual data~\cite{hong2023diminishing}, gene sequences~\cite{zvyagin2023genslms}, and metabolite SMILE strings~\cite{zeng2022deep}. Additionally, we feed the representations generated by the multi-modal encoder into the Graph Convolutional Networks (GCN)-based model~\cite{kipf2016semi} to find equivalent or similar vertices as inter-graph links. To mitigate the influence of ``\textit{danglings}'' in identifying inter-graph links, we perform a ``\textit{dangling}'' elimination strategy. Starting with an initial ``\textit{dangling}'' candidate set based on prior knowledge from the multi-modal encoder, this strategy dynamically adjusts as vertex representations evolve, and progressively removes ``\textit{danglings}'' to minimize their impact. Finally, our interactive knowledge transfer strategy facilitates knowledge exchange between two graphs via inter-graph links. This approach generates new intra-graph links from inter-graph connections, which in turn aids the graph alignment model in identifying additional inter-graph links. By interactively proposing both inter-graph and intra-graph links, we enrich the metabolic structure and enhance association prediction performance.
In the association prediction stage, unlike traditional link prediction tasks that identify a single target gene from a source metabolite and relation, our new task identifies all potential genes associated with a specific metabolite targeted for overproduction. Our \modelname predicts all viable genes linked to given production metabolites and assesses the feasibility of a gene catalyzing a metabolite for other productions, by integrating knowledge from multiple metabolic graphs. We
evaluate our method across three link prediction baselines and show it outperforms the baselines in most experiments, with improvements of up to 12.3\%. Our contributions are threefold:
\begin{itemize}
    \item We simplify the complex task of target gene prediction for metabolite production by framing it as gene-metabolite association prediction, significantly easing the challenge of extracting this specific information from diverse sources.
    \item  We leverage the combined strengths of PLMs and structure encoders within our interactive knowledge transfer framework to enhance the robustness of association predictions.
    \item We construct a graph structure from the metabolisms of two model microorganisms, \textit{Saccharomyces cerevisiae} and \textit{Issatchenkia orientalis}, to support future complex inference tasks such as enzyme prediction and compound overproduction.
\end{itemize}

\section{Methodology}

\begin{figure*}[ht]
  \includegraphics[width=\textwidth]{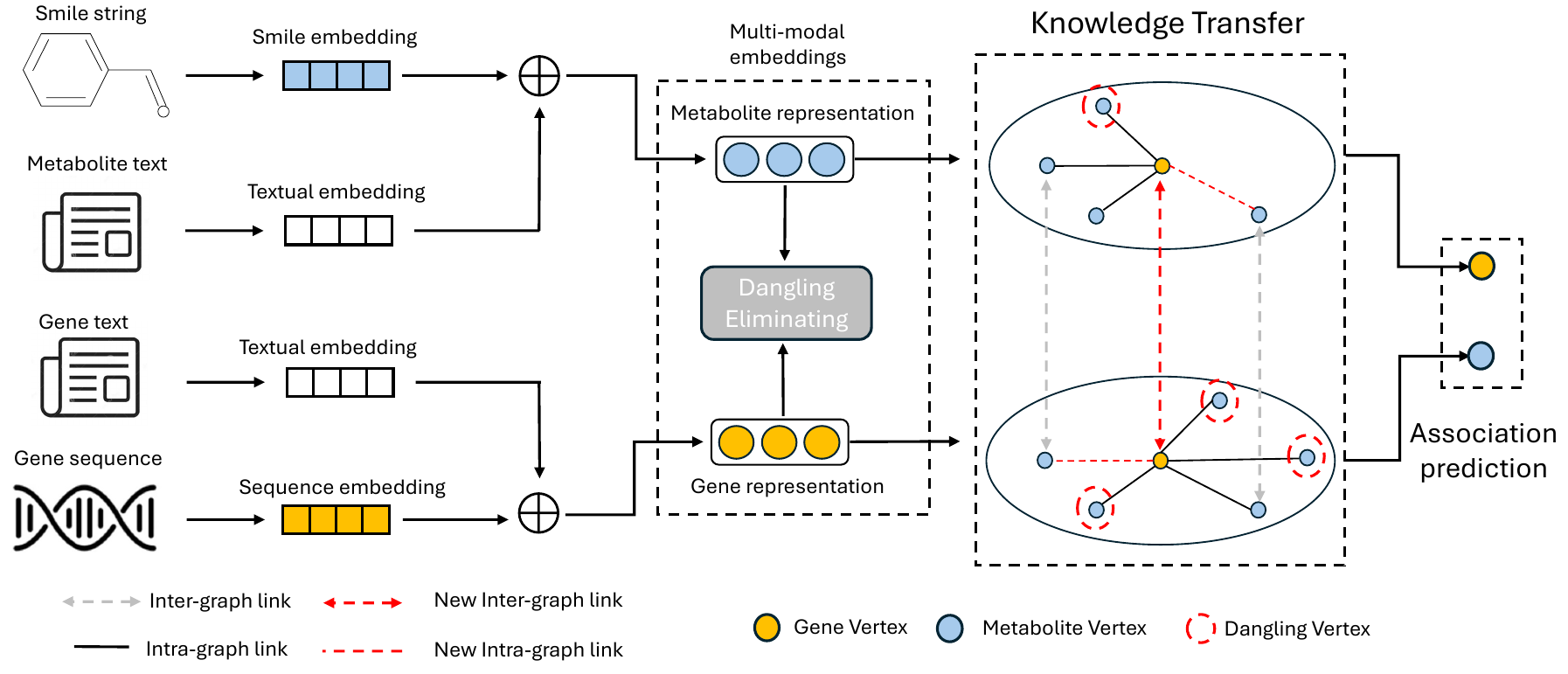}
  \caption{The overview of \modelname framework.}
  \label{fig:framework}
\end{figure*}

\subsection{Problem Definition}
\begin{definition}[Metabolic graph]
A metabolic graph consists of a series of metabolic reactions represented as follows: \[c_1 + \cdots + c_k \xrightarrow{\text{$g_1 + \cdots + g_m$}} c_{k+1} + \cdots + c_n\]
This equation illustrates that the metabolites on the left-hand side, denoted as compounds $c_1$ to $c_k$, are catalyzed by enzymes produced by genes $g_1$ to $g_m$, resulting in the production of metabolites $c_{k+1}$ to $c_n$.
Based on these reactions, we construct a series of triples for the metabolic graph \( M = \{V, D, T\} \): \( C \) and \( G \) are sets of metabolites and genes, respectively, serving as vertices \( V = C \cup G \); \( D \) denotes the direction between genes and metabolites (\textit{left} for reactants and \textit{right} for products); \( T \) is composed of tuples in \( C \times D \times G \), with each tuple \( t = (c, d, g) \) containing \( c \in C \), \( d \in D \), and \( g \in G \).
 For example, from the reaction \texttt{CO4896} $\xrightarrow{\text{HIS6}}$ \texttt{CO4916}, we  derive two triples: $t_1$ = (\texttt{CO4896}, left, \texttt{HIS6}) and $t_2$ = (\texttt{CO4916}, right, \texttt{HIS6}). The metabolic graph $M$ is a heterogeneous graph containing two types of vertices: metabolites ($c$) and genes ($g$), where each type can only establish connections with the other.

\end{definition}

\begin{definition}[Graph alignment]
    Given two distinct graphs from two microorganisms $M_1 = \{C_1 \cup G_1, D_1, T_1\}$ and $M_2 = \{C_2\cup G_2, D_2, T_2\}$, we aim to identify any identical vertices between these graphs, denoted as inter-graph links $\mathcal{E}_{inter} = \{(c_1, c_2) | c_1 \equiv c_2, c_1 \in C_1, c_2 \in C_2\} \cup \{(g_1, g_2) | g_1 \equiv g_2, g_1 \in G_1, g_2 \in G_2\}$. Prior to analysis, we may already have a set of known inter-graph links, known as $\mathcal{E}^*_{inter}$, which are utilized as the training data for the graph alignment task. 
\end{definition}

\begin{definition}[Gene-Metabolite Association Prediction]
    Given a metabolic graph $M = \{C\cup G, D, T\}$, the gene-metabolite association prediction task aims to predict missing triples $t\in T = (c\in C, d\in D, g\in G)$. Formally, given any potential triple $t\in T = (c\in C, d\in D, g\in G)$, the model needs to predict whether such triple is valid.
\end{definition}

\subsection{Framework Overview}
Figure \ref{fig:framework} illustrates our \modelname framework, which comprises two stages:  enriching graph structures with external knowledge and predicting gene-metabolite associations using these enriched graphs. Given a pair of metabolic graphs $M_1, M_2$, we first use a multi-modal encoder to generate initial vertex embedding based on text, SMILE string of metabolites, and gene sequences. Then, we align two metabolic graphs $M_1, M_2$ and enrich their structures through graph alignment, ``\textit{dangling}'' elimination, and interactive knowledge transfer. Finally, we conduct association predictions based on enriched metabolic graphs $M_1, M_2$.

\subsection{PLM-based multi-model encoder}
To fully utilize the external knowledge of all metabolites and genes within the given metabolic graphs, we leverage the strong expressiveness of PLMs for vertex initialization. We incorporate various types of knowledge, integrating them as the prior knowledge for each vertex.

\paragraph{Text embedding}
Text embedding forms the foundation of all multi-modal embedding processes. We select two types of textual context: surface form, which serves as the identifier of different vertices, and description, which reveals several key attributes detailing the structure, function, and role of various genes or metabolites. Given the text sequence including a set of words $l = \{w_1, \dots, w_n\}$, we utilize a pre-trained state-of-the-art ScholarBERT \cite{hong2023diminishing} to obtain the contextualized hidden states. We then apply mean pooling to acquire the final embedding. The final text embedding for each vertex $v_l$ (where $v$ could be a metabolite vertex $c$ or gene vertex $g$) is produced by concatenating the surface form embedding $l_{sur}$ with the description embedding $l_{des}$, expressed as $v_l = [l_{sur}, l_{des}]$.

\paragraph{External knowledge embedding}
As some micro-organisms, such as Issatchenkia orientalis, are still understudied, the surface form information and description of some genes are often unavailable. We further incorporate external knowledge crucial for identifying the vertex pair-wise similarity. We choose the SMILE string for metabolites and the FASTA sequence for genes. Specifically, for molecules, we employ  KV-PLM \cite{zeng2022deep}, pretrained to bridge the molecule structure with biomedical text, to generate molecule SMILE string embedding $c_{smile}$.
For genes, we choose GenSLMs \cite{zvyagin2023genslms}, specially designed for representing gene sequence, to generate gene embedding $g_{seq}$.

\paragraph{Multi-modal embedding fusion}
Each modality-specific embedding characterizes the vertex identities from disparate aspects. Since the $c_{smile}$ and $g_{seq}$ come from different PLMs, they are in different embedding spaces. To use them together, we first transform these embeddings into the same space through a linear projection layer $\Theta$:
\begin{flalign*}
    && \hat{c}_{smile} = \Theta_1 (c_{smile}) &&
    && \hat{g}_{seq} = \Theta_2 (g_{seq}) &&
\end{flalign*}
where $\Theta_1$, $\Theta_2$ are linear projection layers used to transform the embeddings of $c_{smile}$ and $g_{seq}$, respectively.
The final vertex initialization is the combination of the literal embedding and the side information embedding:
\begin{flalign*}
    && \hat{c} = \Theta_3[c_l, \hat{c}_{smile}] &&
    && \hat{g} = \Theta_3[g_l, \hat{g}_{seq}] &&
\end{flalign*}
where $[.]$ means concatenation, $\Theta_3$ is linear projection layer, and the concatenated embedding is linearly transformed by a shared layer.

\subsection{Interactive Knowledge Transfer}
After acquiring the vertex initialization from the multi-modal encoder, we use them for the following interactive knowledge transfer. We utilize graph alignment and association prediction to perform both inter-graph link and intra-graph transfer to boost the final association prediction performance.

\paragraph{Graph alignment}
We employ GCN \cite{kipf2016semi} as the base model to aggregate neighboring information in the metabolic graphs. Specifically, we incorporate a gating mechanism in the aggregation procedure to control noisy neighbors. We implement Highway gate \cite{srivastava2015highway} to balance the information flow between a vertex and its neighbors through  a gated combination of its input and the original output. Assume $v^n$ represents the vertex embedding at $n$-th layer of GCN, the vertex embedding $v^{n+1}$ at $(n+1)$-th layer is:
\begin{equation}
    v^{n+1} = GCN(v^n)
\end{equation}
where $GCN$ represents the neighbor aggregation of the GCN layer. The output embedding $\hat{v}^{n+1}$ from the highway gate is:
\begin{align}
    gate(v^n) &= \sigma(\Theta_4(v^n)) \\
    \hat{v}^{n+1} &= gate(v^n) v^{n+1} + \left(1-gate(v^n)\right) v^n
\end{align}
where $\sigma$ is a sigmoid function, and $\Theta_4$ is a linear projection layer. After the filter process of the highway gate process, we further enhance the influence of important neighboring vertices by using Graph Attention Network (GAT)~\cite{velivckovic2017graph}, which dynamically assigns weight to each neighbor based on their attention scores: 
\begin{equation}
    v^{n+1}_{out} = [v, GAT(\hat{v}^{n+1})]
\end{equation}
where $GAT$ denotes the weighted aggregation of the GAT layer.

We minimize the distance between the two vertices connected by known inter-graph links $\mathcal{E}^*_{inter}$ by the margin-based ranking loss objective funciton\cite{wang2018cross}:
\begin{equation}
    \mathcal{L}_{GA} = \sum_{(v_i, v_j) \in \mathcal{E}^*_{inter}} \sum_{(v'_i, v'_j) \in \mathcal{E'}^*_{inter}} [(d(v_i, v_j) - d(v'_i, v'_j) + \gamma]_+
    \label{eq:ga_loss}
\end{equation}

where $\mathcal{E'}^*_{inter}$ denote the set of randomly sampled negative inter-graph links, $\gamma > 0$ is the margin, $[.]_+ = max(., 0)$, and $d(.)$ computes the Manhattan distance between two vertices.

\paragraph{``Dangling'' eliminating}
To mitigate the negative impact of ``\textit{dangling}'' vertices in the graph alignment task, we propose a ``\textit{dangling}'' eliminating strategy. Leveraging the rich prior knowledge in vertex initializations from the multi-modal encoder, we generate an initial ``\textit{dangling}'' candidate set to protect the following alignment learning. Specifically, a ``\textit{dangling}'' vertex is expected to be far from its surrounding vertices and should have a lower ranking than vertices in the other graph. Therefore, any vertex in the ``\textit{dangling}'' candidate set should guarantee the following conditions:
\begin{equation}
\left\{
\begin{aligned}
&\alpha = \max_{(v_i, v_j) \in \mathcal{E}^*_{inter}} (d(v^0_i, v^0_j))\\
&d(v, NN(v)) > \alpha\\
&\forall v' \in V', R(v | v') > \theta |V'|
\end{aligned}
\right.
\label{eq:dangling}
\end{equation}
where $v^0_i, v^0_j$ denotes the vertex initial representations, which are the muli-modal embeddings, $NN(.)$ denotes the nearest neighbors, and $R(v | v')$ denotes the ranking position of vertex $v$ in relation to vertex $v'$, based on increasing distances among all vertices in the set $V$ to $v'$. During alignment training, we continually update the ``\textit{dangling}'' candidate set to reflect changes in the embedding distances.

For vertices in the dangling candidate set, we gradually reduce their influence by lowering their weight in the adjacency calculations of connected vertices, thereby alleviating heterogeneity.
For example, in figure \ref{fig:walk}, eliminating the dangling vertices \texttt{YDR020C} and \texttt{G\_G514} helps the neighboring vertices (1, 2-hops away) of \texttt{YLR245C} and \texttt{G\_G4580} to exhibit more similarity. This similarity facilitates easier discovery of inter-graph links between these vertices.
When a vertex remains in the ``\textit{dangling}'' set for a specific number of epochs $k$, it is classified as formally ``\textit{dangling}'' and is then permanently excluded from the neighbor lists of its connected vertices. 
This approach ensures that equivalent vertices across graphs develop increasingly similar neighboring contexts, reducing distances between them, which in turn facilitates more accurate discovery of new inter-graph links.

\paragraph{Iteractive Knowledge Transfer}
After establishing the inter-graph links as bridges between metabolic graphs, we perform the knowledge transfer through these bridges. 
The knowledge transfer process encompasses two primary aspects: intra-graph link transfer and inter-graph link transfer. These two forms of knowledge transfer mutually benefit each other, boosting overall effectiveness.

\textit{Inter-graph link transfer.} 
During the alignment training, we iteratively infer new inter-graph links. Considering the hubness and isolation issues \cite{sun2020benchmarking} in high-dimensional space, we propose a simple but effective strategy to select the vertex pairs that are most highly similar. Specifically, for any pair of inferred inter-graph link $(v_i, v_j)$, it should satisfy the following two conditions:
\begin{equation}
\left\{
\begin{aligned}
d(v_i, v_j) < \gamma_d \\
v_j = NN(v_i) && \& && v_i = NN(v_i)
\end{aligned}
\right.
\label{eq:inter-graph}
\end{equation}
Where $\gamma_d$ is the distance threshold for selecting inter-graph links. We limit the choice of inter-graph links by a threshold due to``\textit{dangling}'' vertices. We define a pair of vertices as a new inter-graph link if they are mutually the nearest neighbors to each other and their distance is sufficiently small. We denote the new inter-graph link set as $\hat{\mathcal{E}}^*_{inter}$.

\textit{Intra-graph link transfer.} 
The intra-graph link transfer performed for the association prediction model contains two parts. Firstly, to bridge graphs via inter-graph links, we derive new triples by swapping the vertices connected by these links. However, the association prediction model, primarily focused on inferring missing triples, lacks calibration across the embedding spaces of two metabolic graphs. Given an inter-graph link $(v_i, v_j) \in \mathcal{E}_{inter}$, if there is a triple $(v_i, r_i, v'_i) \in T_i$, we generate a corresponding new triple $(v_j, r_i, v'_i)$. Similarly, for a triple $(v'_i, r_i, v_i) \in T_i$, we generate a new triple $(v'_i, r_i, v_j)$. The same rule applies to the triples of $v_j$ transfer to $v_i$.
The second part involves generating intra-graph links within a single graph. In particular, for each pairs of inter-graph links $(v_i, v_j)$ and $(v'_i, v'_j)$, if there is a relation $r$ connecting $v_i$ and $v'_i$, we will also add the same $r$ to $v_j$ and $v'_j$.
The newly inferred intra-graph links not only augment the association prediction data but also help gradually calibrate the inter-graph links. These links are denoted as $\mathcal{E}^*_{intra}$.

\subsection{Association prediction}
As the training process progressively expands the number of new intra-graph and inter-graph links, we perform the association prediction based on the graph structure enriched by both types of links. We formalize the association prediction task by a triple-level function: $f(t) = |v + r - v'|$, designed to measure the plausibility of a triple $t = (v, r, v')$. Here, $v$ and $v'$ are vertices, and $r$ represents the relationship between them. The triple can be both within-graph or cross-graph. The model learning of association prediction follows the margin-based objective:
\begin{equation}
    \mathcal{L}_{LP} =  \sum_{t \in T + \mathcal{E}^*_{intra}} \sum_{t' \in T'} [f(t) - f(t') + \beta]_+
    \label{eq:lp}
\end{equation}
where $T'$ comprises the negative triples set generated by random sampling one of the vertices from metabolic graph, $\beta > 0$ is the margin maintained in the association prediction task.

\section{Experiment}

\subsection{Dataset setup}
Table \ref{table: dataset} shows the statistics for the genome-scale metabolic models of two micro-organisms, \textit{Saccharomyces cerevisiae} (SC) \cite{luConsensusCerevisiaeMetabolic2019b} and \textit{Issatchenkia orientalis} (IO) \cite{suthersGenomescaleMetabolicReconstruction2020}. 
SC is the most commonly used model organism, while IO is a non-model organism known for its advantageous properties.
As the metabolic information for IO is pre-annotated based on SC data, these two microorganisms share numerous genes and metabolites, which are identifiable as inter-graph links.
We collect information on metabolic mechanisms from available sources\footnote{\url{https://github.com/SysBioChalmers/yeast-GEM}}\footnote{\url{https://github.com/maranasgroup/iIsor_memote}}.
This includes lists of genes and metabolites, biological reaction data, and references from official knowledge bases. Additionally, we extract multi-modal features such as textual content, SMILES strings, and gene sequences from existing knowledge bases\footnote{\url{https://www.uniprot.org/}}\footnote{\url{https://www.yeastgenome.org/}}.
For the gene-metabolite association prediction task, we split the intra-graph links of the metabolic graph into 60\% for training, 30\% for testing, and 10\% for validation.
In the graph alignment task, we forego traditional training data due to the unreliability and incompleteness of gene inter-graph links; instead, we rely on supervision signals derived from initial multi-modal embeddings.
 Table \ref{tab:split} shows the distribution of genes and metabolites, as well as the training, testing, and validation splits in our dataset.

\begin{table}[!htb]
\centering
% \scriptsize
	\centering
	% \begin{tabular}{lccccc}
\caption{Statistics of metabolic graphs of SC and IO, where Intra denotes intra-graph links and Inter denotes inter-graph links.
}
\label{table: dataset}
\begin{tabularx}{\linewidth}{>{\hsize=1.4\hsize}X>{\centering\arraybackslash\hsize=1\hsize}X>{\centering\arraybackslash\hsize=1\hsize}X>{\centering\arraybackslash\hsize=0.8\hsize}X>{\centering\arraybackslash\hsize=0.8\hsize}X}
\toprule
 \textbf{ Microorganism} &\textbf{\#Gene} &\textbf{\#Metabolite}&\textbf{Intra} &\textbf{Inter}\\%&\textbf{inter G} &\textbf{inter M}  \\
 \midrule
SC              & 1,097 & 1,046& 7,746 & 781 \\%& 613 & 781\\
IO           & 850 & 1428 & 5,943 & 781 \\%613 & 781\\
% \hdashline
% YI          & 23.96  &29.82 & 27.65 &31.48  & 32.76\\
% RT	             &0.99	&1.43	&2.39	&1.61	&2.45\\
\bottomrule
\end{tabularx}
	% \end{tabular}
\end{table}

\begin{table}[!t]
	\centering
	\caption{The gene and metabolite distribution for train, test, valid data.}
	\resizebox{0.95\linewidth}{!}{
		\begin{tabular}{lcclccl}
			\toprule
			\multirow{2}{*}{\textbf{Type}} & \multicolumn{3}{c}{\textbf{SC}} & \multicolumn{3}{c}{\textbf{IO}} \\
			\cmidrule(lr){2-4} \cmidrule(lr){5-7} 
			& \textbf{Train} & \textbf{Test} & \textbf{Valid} & \textbf{Train} & \textbf{Test} & \textbf{Valid} \\ \midrule
			 Gene & 1,097 & 1,097 & 1,077 & 850 & 850 & 831 \\
			 Metabolite & 1,046 & 1,046 & 1,011 & 1,428 & 1,426 & 1,248 \\
			\bottomrule
	\end{tabular}}
	\label{tab:split}
\end{table}

\subsection{Implementation details}
The framework is implemented based on PyTorch\footnote{\url{https://pytorch.org/}} and Huggingface Transformer\cite{wolf-etal-2020-transformers}. For the association prediction,  we corrupted each correct test triple each correct test triple $(g, d, c)$ by replacing the $g$ with another $g'$ or replacing the $c$ with another $c'$. 
We set the negative sampling rate at 10 and the learning rate at $5e^{-4}$. 
During training, we use the validation set to select the best model.
Our new task differs from traditional link prediction, which provides the subject vertices and relations;  in our case, the model lacks prior knowledge of the candidate target objects.
We utilize the classic evaluation metrics, which are precision, recall, and F1. 
During testing,  the model was presented with both correct and corrupted triples and tasked with predicting their validity.
We set the multi-modal embedding dimension to 512 for both the association prediction model and the graph alignment model. 
For the graph alignment task using a GCN-based model, we perform a grid search to select the optimal number of hidden layers $(1, 2, 3)$, initial Adam learning rates of $1e^{-4}, 4e^{-4}, 1e^{-1}$, and set the margin $\gamma$ in Equation \ref{eq:ga_loss} to 5. Additionally, we set $\gamma_d = 0.4$ in Equation \ref{eq:inter-graph}. 
In equation \ref{eq:dangling},  we set $\alpha$ to $0.7$ and $\theta$ to $0.8$.
During association prediction training, $\beta$ to $1$ for Equation \ref{eq:lp}.

% The vertice embedding dimension is 500.
% As link prediction tasks focus on the learning of pure structure, incorporating prior knowledge does not bring improvement, the embeddings for this task is random initialized with dimension 500.
\subsection{Main results}
Table \ref{tab:main} illustrates the association prediction results for \modelname and other strong baselines on SC and IO. 
Since our framework adapts to various structure encoders, we test its performance alongside popular link prediction models, including TransE~\cite{bordes2013translating}, RotatE~\cite{sun2018rotate}, and DisMult ~\cite{yang2014embedding}. RotatE excels in modeling various relational patterns, while DisMult employs a diagonal assumption for relations.
Overall, these three models incorporating our model can achieve an improvement from 8\% to 12\%. 
DisMult performs worse than TransE and RotatE in both settings, with or without \modelname, due to its inability to model complex graph patterns effectivelly.
RotatE and TransE perform similarly in both settings. 
% Language models all fail to achieve satisfactory performance. This is because language models are not proficient in modeling the triple-based structure, while Trans-based models can directly formalize the semantic information for link prediction. 
% Our model \modelname outperforms all baselines and achieves an 11\% gain compared with the 2ed best baseline. 
The improvement in our model's performance is primarily attributed to its interactive knowledge transfer, which leverages prior knowledge to capture consistencies between SC and IO, thus enriching the metabolic graph structure and enhancing association prediction outcomes.
% The main reason for our model's improvement is its assistance with interactive knowledge transfer.
% Prior knowledge captures consistency between SC and IO and enriches the metabolic graph structure, enhancing the final association prediction performance.

\begin{table}[!htb]
% \vspace{-1em}
\centering
\small
\caption{The performance in test set of SC and IO, where the best and second best performance is in bold and italic respectively.\label{tab:main}.}
\begin{tabularx}{\linewidth}{>{\hsize=4\hsize}X>{\centering\arraybackslash\hsize=0.5\hsize}X>{\centering\arraybackslash\hsize=0.5\hsize}X>{\centering\arraybackslash\hsize=0.5\hsize}X>{\centering\arraybackslash\hsize=0.5\hsize}X>{\centering\arraybackslash\hsize=0.5\hsize}X>
{\centering\arraybackslash\hsize=0.5\hsize}X}
\toprule
\multirow{ 2}{*}{\textbf{Model}}&\multicolumn{3}{c}{\textbf{SC}}&\multicolumn{3}{c}{\textbf{IO}}\\
&\textbf{P}&\textbf{R}&\textbf{F1}&\textbf{P}&\textbf{R}&\textbf{F1}\\
\midrule

RotatE &0.492 & 0.682 & 0.572 & 0.360 & 0.507 & 0.421\\
DisMult &0.456 & 0.682 & 0.546 & 0.304 & 0.522 & 0.384\\ 
TransE             & 0.749           & 0.459          & 0.569          & 0.463          & 0.395 & 0.426\\\hdashline
% SimKGC           & 0.197 & 0.531 & 0.287 & 0.181 & 0.475 &  0.262\\
% LP4GP (w/o kt, mm, be)                   & 0.749           & 0.459          & 0.569          & 0.463          & 0.395 & 0.426\\

RotatE + \modelname                   & \textbf{0.826}           & \textbf{0.601}         & \textbf{0.695}         & \textit{0.396}        & \textit{0.680} & \textit{0.501}\\
DisMult + \modelname                   & 0.643          & 0.677          & 0.660          & 0.367          & 0.684 & 0.478\\
TransE + \modelname                   & \textit{0.740}          & \textit{0.636}          & \textit{0.684}         & \textbf{0.583}          & \textbf{0.455} & \textbf{0.511}\\
% LP4GP                   & 0.740          & 0.636          & 0.684         & 0.583          & 0.455 & 0.511\\
\bottomrule
\end{tabularx}
\end{table} 

% \begin{table}[!htb]
% \centering
% % \scriptsize
% 	\centering
% 	% \begin{tabular}{lccccc}
% \begin{tabularx}{1.0\linewidth}{>{\hsize=2.15\hsize}X>{\centering\arraybackslash\hsize=0.77\hsize}X>{\centering\arraybackslash\hsize=0.77\hsize}X>{\centering\arraybackslash\hsize=0.77\hsize}X>{\centering\arraybackslash\hsize=0.77\hsize}X>
% {\centering\arraybackslash\hsize=0.77\hsize}X>{\centering\arraybackslash\hsize=0.77\hsize}X}
% \toprule
%  \textbf{In-graph(SC)} &\textbf{Dev P} &\textbf{Dev R}&\textbf{Dev F1} &\textbf{Test P}&\textbf{Test R} &\textbf{Test F1} \\
%  \midrule
% % BiomedGPT(no lit)    & 0.133  &0.964 & 0.234 &0.132  & 0.955 & 0.232\\
% % BiomedGPT(with lit) & 0.200  &0.396 & 0.266 &0.206  & 0.453 & 0.283\\
% TransE              & 0.707 & 0.495& 0.582 & 0.749           & 0.459          & 0.569\\
% RotaE & \\
% DisMult &\\ \hdashline
% SimKGC           & 0.197 & 0.531 & 0.287 & 0.181 & 0.475 &  0.262\\
% \hdashline
% \modelname & 0.767 & 0.652 & 0.705 & 0.740 & 0.636 & 0.684\\
% \bottomrule
% \end{tabularx}
% 	% \end{tabular}
% \caption{P, R, F1 scores for SC with different baseline models.
% }
% \label{tab: main_result}
% \end{table}

\subsection{Ablation Study}
To evaluate the effectiveness of each proposed component, we test \modelname by removing each component in turn. Since TransE is more lightweight, we use it for ablation study. Table \ref{tab:KT} shows the results. We include ScholarBERT+Classifier as an additional surface form-only baseline. 
We denote the performance of removing each component as ``w/o [component]'', where the components include knowledge transfer (kt), multi-model embeddings (mm), and ``\textit{dangling}'' eliminating (de).

\noindent \textbf{Impact of interactive Knowledge transfer}
 The multi-modal embeddings and the ``\textit{dangling}'' eliminating components are used to build inter-graph links for knowledge transfer.
 Therefore, when removing the knowledge transfer component, we also need to remove the other two components.
 Without these three components, our model reverts to the TransE baseline. 
 As it accesses only single microorganism information, the graph structure cannot be enriched with knowledge from another microorganism. Consequently, performance matches the TransE baseline, which our association prediction model follows. However, with interactive knowledge transfer, the model significantly enriches the graph structure from both metabolic graphs, leading to a substantial enhancement in overall performance.

\begin{table}[!htb]
% \vspace{-1em}
\centering
\small
\caption{The ablation study of TransE-based \modelname, where kt, mm, de refer to knowledge transfer, multi-modal embeddings, and dangling eliminating respectively. Best performance is in bold font\label{tab:KT}. }
\begin{tabularx}{\linewidth}{>{\hsize=4\hsize}X>{\centering\arraybackslash\hsize=0.5\hsize}X>{\centering\arraybackslash\hsize=0.5\hsize}X>{\centering\arraybackslash\hsize=0.5\hsize}X>{\centering\arraybackslash\hsize=0.5\hsize}X>{\centering\arraybackslash\hsize=0.5\hsize}X>
{\centering\arraybackslash\hsize=0.5\hsize}X}
\toprule
\multirow{ 2}{*}{\textbf{Model}}&\multicolumn{3}{c}{\textbf{SC}}&\multicolumn{3}{c}{\textbf{IO}}\\
&\textbf{P}&\textbf{R}&\textbf{F1}&\textbf{P}&\textbf{R}&\textbf{F1}\\
\midrule
% TransE              & 0.707 & 0.495& 0.582 & 0.708 & 0.480 & 0.572\\
% RotaE &0.492 & 0.682 & 0.572 & 0.360 & 0.507 & 0.421\\
% DisMult &0.456 & 0.682 & 0.546 & 0.304 & 0.522 & 0.384\\  \hdashline
% SimKGC           & 0.197 & 0.531 & 0.287 & 0.181 & 0.475 &  0.262\\
ScholarBERT Classifier & 0.584 & 0.532 & 0.557 & 0.421 & 0.487 & 0.451\\
\modelname w/o kt, mm, de                   & 0.749           & 0.459          & 0.569          & 0.463          & 0.395 & 0.426\\
\modelname w/o mm, de                   & 0.567           & 0.683         & 0.620        & 0.416         & 0.475 & 0.443\\
\modelname w/o mm                   & 0.617           & 0.680         & 0.648         & 0.449         & 0.468 & 0.458\\
\modelname w/o de                   & 0.672           & 0.677          & 0.675          & 0.660          & 0.395 & 0.494\\
\modelname                   & \textbf{0.740}          & \textbf{0.636}          & \textbf{0.684}         & \textbf{0.583}          & \textbf{0.455} & \textbf{0.511}\\
\bottomrule
\end{tabularx}
\end{table} 

\noindent \textbf{Impact of Multi-modal embeddings}
To test the advantages that language models bring for detecting inter-graph links, we replace the multi-modal embeddings with random initialization in our framework and keep other parts unchanged.
Comparing our full model with the one removing multi-modal embeddings, the F1 performance suffers a decrease from 4\% to 6\%.
The reason for the drop is the heterogeneous structure between the metabolic graphs of SC and IO.
Even though GCN can capture some consistency between graphs and detect some inter-graph links, relying solely on structural information proves very challenging compared with incorporating the prior knowledge pre-trained by a large number of materials from different PLMs.
As the number of inter-graph links decreases, the graphs fail to acquire sufficient enrichment, resulting in a performance drop.
When the dangling elimination strategy is unavailable, comparing models without it (de) to those without multi-modal (mm), we observe about a 5\% F1 score drop.
This phenomenon verifies the importance of leveraging knowledge from pretrained language models.

\noindent \textbf{Impact of dangling eliminating strategy}
To assess the improvement from the dangling elimination strategy, we evaluate the performance without detecting the ``danglings'' and directly perform the alignment learning. 
From table \ref{tab:KT}, we can see the performance also suffers a small decrease with around 2\% when removing this strategy from the full model.
This is because, without this process, the difficulty of detecting inter-graph links increases.
Due to the existence of ``\textit{dangling}'' vertices, the vertice pairs that should be linked cross graphs failed to be detected with similar structures, resulting in a decreased amount of inter-graph links. 
Therefore, without a sufficient amount of inter-graph links, the graph structure cannot be fully enriched, leading to poorer outcomes. 
We can also notice that when the multi-modal component is replaced by random initialization, the improvement brought by the dangling eliminating strategy increases to about 3\%. 
This is because, with multi-modal embeddings unavailable, the discovery of inter-graph links is solely based on structure content. 
With multi-modal embeddings providing prior knowledge, the structural impact on finding inter-graph links diminishes.

% \noindent \textbf{Impact of different inter-graph links ratio}
% todo: plan to test different inter-graph link ratios in different cases. 
% eg: when no multi-modal embeddings, the impact of dangling detection
% test setting different threshold for inter-graph links

\noindent \textbf{Complementarity between language model and structure encoder} 
To test if combining language models with structure encoders can improve performance, we use ScholarBERT embeddings. These are concatenated from gene-metabolite association triples as input for a binary classifier.
With only a single graph information, we can see that SC's performance is similar to our model's, while its IO result achieves a 2.5\% improvement. 
This is because the IO metabolic graph is sparse, making the single graph structure inadequate for association prediction.
When we combine the language models with the structure encoders, our approach aligns with the ``mm'' baselines.
We observe that using the language model to generate inter-graph links significantly enriches the metabolic graph. This enhancement leads to an F1 score improvement of over 10\%.

\subsection{Case study}
% todo: add cases that our model can discover but data not contain, but it is true link
% infer intra-graph link by inter-graph links
% case 1 for SC: YPL214C, carbon dioxide
% case 2 for IO:

In this section, we conduct case studies to further demonstrate the effectiveness of our \modelname, i.e., how the interactive knowledge transfer can be used to fix the incorrect prediction cases.

1) For the first two lines of table \ref{table: case}, without the interactive knowledge transfer, which means we perform the gene-metabolite association (G-M) prediction based on single metabolic graph information, the model failed to identify the given G-M link as correct for the IO graph. 
While the SC contains more complete metabolism information, the association prediction model of SC has higher performance.
Thus, the SC's association prediction model successfully identifies this G-M association, as its confidence exceeds 0.5.

When we add the knowledge transfer to integrate the information from both graphs, we can see the structure similarity between \texttt{G\_JL09\_g616} and \texttt{YDR483W} (genes) is very high, and the structure similarity of \texttt{M\_akg\_c} and \texttt{akg\_c} (metabolite) is also high. 
Although literal gene information is unavailable for \texttt{G\_JL09\_g616}, their gene sequences are highly similar (0.786).
Additionally, it is clear that \texttt{M\_akg\_c} and \texttt{akg\_c} refer to the same metabolite.
Based on our knowledge transfer rule, we directly transfer the link (\texttt{YDR483W}, left, \texttt{akg\_c}) to (\texttt{G\_JL09\_g616}, left, \texttt{M\_akg\_c}). 
After the knowledge transfer, our model successfully identifies this correct link.

2) In this case, our model without the knowledge transfer failed to identify this correct G-M link again. 
Although it is easy to identify that \texttt{M\_nadph\_m} and \texttt{nadph\_m} are the same metabolite, identifying the equivalence between gene \texttt{G\_JL09\_g1308} and \texttt{YER073W} is very difficult.
The structure similarity between the two genes is low, as these genes have a large ratio of different reactions in IO and SC, respectively. 
Manual evaluations reveal that \texttt{G\_JL09\_g1308} is involved in almost twice as many reactions in IO as \texttt{YER073W} in SC.
Additionally, although their gene sequence similarity is high (0.695), this evidence is not convincing due to significant structural discrepancies. 
The dangling elimination strategy mitigates the impact of dangling vertices and increases structural similarity, aiding the model in recognizing that these genes are the same.
Based on the knowledge transfer rule, as we already justify the correctness of G-M link (\texttt{YER073W}, right, \texttt{nadph\_m}), and we identify the inter-graph link (\texttt{G\_JL09\_g1308}, \texttt{YER073W}) and (\texttt{M\_nadph\_m}, \texttt{nadph\_m}), we can successfully infer the new intra-graph link (\texttt{G\_JL09\_g1308}, right, \texttt{M\_nadph\_m}).
\begin{table*}[!htb]
\centering
% \scriptsize
	\centering
	% \begin{tabular}{lccccc}
\caption{Quantitive analysis of G-M links.
}
\label{table: case}
\begin{tabularx}{1.0\linewidth}{>{\hsize=2.15\hsize}X>{\centering\arraybackslash\hsize=0.77\hsize}X>{\centering\arraybackslash\hsize=0.77\hsize}X>{\centering\arraybackslash\hsize=0.77\hsize}X>{\centering\arraybackslash\hsize=0.77\hsize}X>
{\centering\arraybackslash\hsize=0.77\hsize}X>{\centering\arraybackslash\hsize=0.77\hsize}X>
% {\centering\arraybackslash\hsize=0.77\hsize}X>
% {\centering\arraybackslash\hsize=0.77\hsize}X>
{\centering\arraybackslash\hsize=0.77\hsize}X}
\toprule
 \textbf{G-M link} &\textbf{G-M prediction(w/o kt)} &\textbf{Gene sim(w/o de)}&\textbf{Metabolite sim(w/o de)}&\textbf{Gene sim}&\textbf{Metabolite sim}
 % &\textbf{Gene literal}&\textbf{Meta literal} &\textbf{Gene sequence}&\textbf{Smile string} 
 &\textbf{G-M prediction}  \\
 \midrule
G\_JL09\_g616,left, M\_akg\_c & 0.443 & 0.912 & 0.835 & 0.978 & 0.893 
% & 0.103 & 0.985 & 0.786 & 1 
& 0.653\\
YDR483W, left, akg\_c        & 0.675 & 0.912 & 0.835 & 0.978 & 0.893 
% & 0.103 & 0.985 & 0.786 & 1 
&  0.784\\
\hdashline
G\_JL09\_g1308, right, M\_nadph\_m          & 0.346  & 0.466 & 0.635 & 0.631 & 0.734 
% & 0.082 &0.974  & 0.695 & 1 
& 0.521\\
YER073W, right, nadph\_m	             & 0.794	& 0.466 & 0.635 & 0.631 & 0.734	
% & 0.082	&0.974	&0.695 & 1
& 0.801\\
\bottomrule
\end{tabularx}
	% \end{tabular}
\end{table*}

\begin{figure}[!htb]
    \centering
    
  \includegraphics[width=1.0\columnwidth]{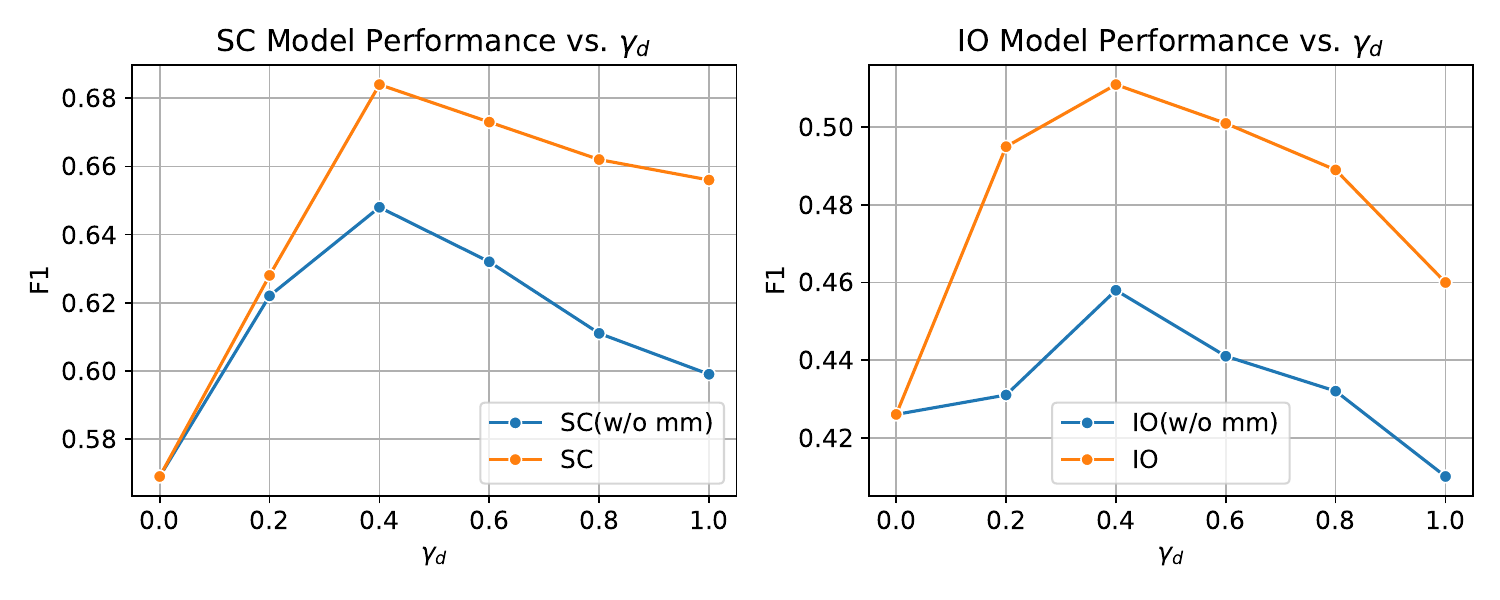}
    \caption{Association prediction performance with different value of $\gamma_d$.}
    \label{fig: both_images}
\end{figure}

3) To assess the impact of varying $\gamma_d$, which selects inter-graph links, we evaluate the performance of both metabolic graphs across different $\gamma_d$ values.
The F1-score of \modelname at a different value of $\gamma_d$ is shown in Figure \ref{fig: both_images}.
To verify the impact of multi-model embeddings, we also test the performance when removing this component.

From the images in Figure \ref{fig: both_images}, we can see the F1-score of our full model outperforms that achieved by replacing multi-modal embeddings with random initialization.
When the value of $\gamma_d$ is 0, we do not leverage any inter-graph links. 
Thus, performance downgrades to that of a single graph.
When increasing $\gamma_d$, the model starts to leverage inter-graph links to generate new intra-graph links. Therefore, the performance increases. However, when the value of $\gamma_d$ is too large, the inferred new inter-graph link may contain noise. So, the performance starts to drop when the leveraged inter-graph links contain incorrect links. 
We notice that when the model includes multi-modal embeddings, the negative impact of setting a high $\gamma_d$ value is less severe than without multi-modal embeddings.
This is because the multi-modal embeddings already have a reasonable initial distance between different vertices.
Therefore, even though a high $\gamma_d$ value allows many inter-graph links, the mutual nearest neighbors limit, and the prior knowledge in the vertex embeddings prevent excessive noisy labels from negatively impacting performance.

% \begin{figure}[htp]
    % \begin{minipage}{0.45\textwidth}
    %     \centering
    %     \includegraphics[width=\textwidth]{figures/sc_plot.pdf} 
    %     \captionsetup{justification=centering}
    %     % \caption{(a)F1 of SC.}
    %     % \label{fig:left_image}
    % \end{minipage}\hfill
    % \begin{minipage}{0.45\textwidth}
    %     \centering
    %     \includegraphics[width=\textwidth]{figures/io_plot.pdf}
    %     \captionsetup{justification=centering}
    %     % \caption{(b)F1 of IO.}
    %     % \label{fig:right_image}
    % \end{minipage}
%     \centering
%     \begin{subfigure}[b]{0.5\textwidth} % Width of subfigure, adjust as needed
%         \centering
%         \includegraphics[width=\textwidth]{figures/sc_plot.pdf} % Include the image
%         \caption{First Subfigure}
%         \label{fig:sub1}
%     \end{subfigure}
%     \vspace{1cm} % Space between the subfigures

%     \begin{subfigure}[b]{0.5\textwidth}
%         \centering
%         \includegraphics[width=\textwidth]{figures/io_plot.pdf}
%         \caption{Second Subfigure}
%         \label{fig:sub2}
%     \end{subfigure}

%     \caption{Example of vertical subfigures}
%     \label{fig:test}
% \end{figure}

\section{Related Work}
% \qingyun{need to add why this task is new}
\subsection{Gene-Protein-Reaction Association Prediction}
% todo: junyu modify
Gene-protein-reaction (GPR) association prediction remains a central task in bioinformatics, aiming to identify gene structures within DNA sequences\cite{machado2016stoichiometric, du2022single}. 
Traditional approaches often relied on manual curation and querying of the structural database, which predicts genes based solely on intrinsic DNA sequence data by exploiting known gene-like features and statistical models such as BLAST\cite{altschul1990basic}, Diamond\cite{buchfink2015fast}, and HMMER\cite{finn2011hmmer}. 
With the advent of next-generation sequencing technologies, gene prediction has increasingly incorporated comparative genomics and machine learning techniques, leveraging both evolutionary conservation and functional genomics data. 
Machine learning methods, particularly those employing deep learning, have shown significant promise in improving the accuracy of gene predictions, capable of modeling complex patterns and dependencies within genomic data that traditional statistical methods might miss. Notably, DeepEC\cite{ryu2019deep}, based on convolutional neural networks (CNNs), has been the state-of-the-art method for predicting enzyme commission numbers.
The recent advancement of CLEAN, which is based on a contrastive learning method, has achieved higher performance than the traditional method.
\cite{yu2023enzyme} 
Once the gene has been annotated with its function, it will be easy to retrieve its role in metabolite catalysis. These gene-protein-reaction (GPR) association prediction methods can uncover the gene's unknown annotations and have great potential to bridge gaps and expand the limited metabolic network for researchers to better comprehend the underlying mechanisms.

\subsection{Graph Embedding}
Graph embedding translates graph-structured data into a low-dimensional space, preserving node connectivity, edge attributes, and complex structures like subgraphs.
It includes Knowledge Graph Embedding and Network Embedding, which convert graphs into vectors.

Network embedding deals with the conversion of network data into vector space. 
Techniques like DeepWalk\cite{perozzi2014deepwalk}, node2vec\cite{grover2016node2vec}, and Struc2vec \cite{ribeiro2017struc2vec} initially used random walks to capture local topology and preserve network neighborhoods. Recent developments \cite{zhang2024network, deng2024noise} incorporate graph neural networks (GNNs) such as Graph Convolutional Networks (GCNs)\cite{kipf2016semi} and Graph Attention Networks (GATs)\cite{velivckovic2017graph,wang-etal-2019-paperrobot}, enhancing the understanding of complex biological networks\cite{su2020network}.
% Initially, methods like DeepWalk\cite{perozzi2014deepwalk} and node2vec\cite{grover2016node2vec} popularized the concept of using random walks\cite{karlin1959random} to capture the local topology of a network, embedding nodes in a way that preserves network neighborhoods. Struc2vec \cite{ribeiro2017struc2vec} started to learn node representations via structural identity and allows for learning representations in separate networks, which firstly provides the possibility to leverage cross-network information. More recent advancements \cite{zhang2024network, deng2024noise} have integrated more sophisticated models, including those based on graph neural networks (GNNs), which directly leverage node feature information and edge connectivity. Techniques such as Graph Convolutional Networks (GCNs)\cite{kipf2016semi} and Graph Attention Networks (GATs)\cite{velivckovic2017graph} have further refined the capability to handle heterogeneity and complex patterns in the data, making them especially suited for biological networks where nodes and edges can represent various biological entities and interactions \cite{su2020network}. These methods have not only provided more accurate embedding but have also enabled a deeper understanding of the underlying structure and function of biological systems.
Knowledge graph embedding transforms entities and relationships into vectors, using translational methods\cite{bordes2013translating, lin2015learning, wang2014knowledge}, factorization\cite{balavzevic2019tucker, kazemi2018simple, wang2018multi}, and neural models\cite{dettmers2018convolutional, shi2017proje, vashishth2019composition, liu2019context, liu2021temporal}. It differs from network embedding, which focuses on node interactions without explicit labels, making it suitable for detecting patterns and community structures in networks.
% In our work, we apply both methodologies to predict links in a metabolism network, leveraging structural and semantic properties to enhance gene prediction accuracy.
In our task, the metabolism network is not rich in semantics but the edges carry the information between vertices, and our gene prediction is formalized as association prediction. Therefore, our methodology design should not only capture the structural and neighborhood properties but also encode the rich semantics information. So we leverage different model designs to realize these two factors.

% Knowledge graph embedding focuses on transforming the entities and relationships in a knowledge graph into a low-dimensional space. It can be generally classified into 3 types. Translational methods interpret a relation as the translation vector from its subject entity to object entity \cite{bordes2013translating, lin2015learning, wang2014knowledge}. Factorization-based methods encode a KG as a binary tensor and decompose it into a set of matrices to represent entities and relations \cite{balavzevic2019tucker, kazemi2018simple, wang2018multi}. Neural models generate KG embedding by predicting entity or relation \cite{dettmers2018convolutional, shi2017proje, vashishth2019composition}. 

% Knowledge graph embedding generally deals with graphs rich in semantics with typed edges and often more structured, whereas network embedding is typically about the interaction between nodes and might not carry explicit labels. So the former is ideal for reasoning and inference over relational data, while the latter is suited for uncovering hidden patterns, community structures, or influential nodes in networks. In our task, the metabolism network is not rich in semantics but the edges carry the information between vertices, and our gene prediction is formalized as association prediction. Therefore, our methodology design should not only capture the structural and neighborhood properties but also encode the rich semantics information. So we leverage different model designs to realize these two factors.

\subsection{Joint graph alignment and link prediction}
% Joint graph alignment and completion is an advanced area of study that focuses on enhancing the predictive capabilities of graph-based models by simultaneously aligning and completing multiple graphs. This approach is particularly relevant in domains where multiple related but distinct graphs are available, such as in multi-species biological networks or social networks from different regions. The core idea is to find a common substructure or alignment among these graphs while also predicting missing nodes or edges, thereby completing the graphs. Techniques developed in this field often utilize cross-graph node similarities and inter-graph edge predictions, leveraging both structural and feature-based similarities to enhance alignment and completion tasks. Recent advancements often employ variational autoencoders and coupled matrix-tensor factorization methods, which have shown considerable success in not only improving the accuracy of the graph alignment but also in effectively inferring missing information. These methodologies have been crucial in understanding complex patterns and interactions within biological data, offering insights that are not readily observable from isolated or incomplete networks. Such dual tasks help in better characterizing gene interactions across species and can significantly impact comparative genomics by providing a more holistic view of gene functionality and evolution.

Joint graph alignment and completion is an advanced area of study that focuses on enhancing the predictive capabilities of graph-based models by simultaneously aligning and completing multiple graphs. 
Previously, multilingual Knowledge Graph Completion (KGC) aimed to enrich KGC by utilizing multiple knowledge graphs (KGs) in various languages, a task quite similar to joint graph alignment and completion.
Initially, the MTransE model \cite{chen2016multilingual} pioneered the extension of knowledge graph embedding from a single language to a multilingual context, facilitating the transfer of information across different language-specific KGs. Subsequent research primarily concentrated on the entity alignment task among diverse KGs \cite{zhang2019multi, sun2020knowledge, guo2022understanding, xin2022ensemble, xin2022informed, xin2022large, liu2023tea}.
The KEnS framework, introduced by \cite{chen2020multilingual}, further enhances monolingual KGC by effectively harnessing the complementary knowledge from multilingual KGs.
AlignKGC \cite{singh2021multilingual} combines KGC with entity and relation alignment within multilingual KGs, thereby improving both KGC accuracy and alignment metrics. Furthermore, SS-AGA \cite{huang2022multilingual} advances multilingual KGC by employing a relation-aware graph neural network that dynamically generates additional potential alignment pairs. Despite these advancements, entity alignment remains the predominant focus in multilingual KGC. Moreover, few of them pay attention to the negative impact of ``dangling'' nodes in the joint learning framework.
% Moreover, the knowledge models in these studies are generally based on monolingual KGC methods, with few addressing the direct transfer of knowledge across languages. 
 In our case, we focus on the link prediction task for gene prediction, our ultimate goal, and have designed a "dangling" elimination algorithm to reduce their influence and improve consistency identification between graphs.

\section{Conclusion}
In this work, we propose a new task, Gene-Metabolite Association Prediction based on metabolic graphs, to automate the process of candidate gene discovery, which is the initial step for the target gene prediction in metabolic engineering. We design a new framework to improve the association prediction performance by enriching the metabolic graph structure. The enrichment process is conducted by integrating knowledge from different metabolic graphs. We resort to the power of PLMs to provide prior knowledge for inter-graph link discovery. 
Then, inter-graph links serve as bridges to integrate and enrich the metabolic graphs.
Finally, the association prediction is performed on the enriched graph, leading to a large improvement in performance.
In the future, we will consider incorporating reaction information in our constructed metabolic graphs.

% \subsection{Subsection Heading Here}
% Subsection text here.

% % needed in second column of first page if using \IEEEpubid
% %\IEEEpubidadjcol

% \subsubsection{Subsubsection Heading Here}
% Subsubsection text here.

% \appendices
% \section{Proof of the First Zonklar Equation}
% Appendix one text goes here.

% % you can choose not to have a title for an appendix
% % if you want by leaving the argument blank
% \section{}
% Appendix two text goes here.

% use section* for acknowledgment
\section*{Acknowledgment}
This work is supported by DOE Center for Advanced Bioenergy and Bioproducts Innovation U.S. Department of Energy, Office of Science, Office of Biological and Environmental Research under Award Number DESC0018420. The views and conclusions contained herein are those of the authors and should not be interpreted as necessarily representing the official policies, either expressed or implied of, the National Science Foundation, the U.S. Department of Energy, and the U.S. Government. The U.S. Government is authorized to reproduce and distribute reprints for governmental purposes notwithstanding any copyright annotation therein.
\bibliographystyle{IEEEtran}
\bibliography{bibtex/bib/IEEEexample}
\end{document}